\documentclass{ecai} 

\usepackage{latexsym}
\usepackage{amssymb}
\usepackage{amsmath}
\usepackage{amsthm}
\usepackage{booktabs}
\usepackage{enumitem}
\usepackage{graphicx}
\usepackage{color}

\newcommand{\BibTeX}{B\kern-.05em{\sc i\kern-.025em b}\kern-.08em\TeX}

\begin{document}


\begin{frontmatter}


\paperid{123} 



\title{Explainable Artificial Intelligence for identifying profitability predictors in Financial Statements} 


\author[A]{\fnms{Marco}~\snm{Piazza}
\thanks{Corresponding Author. Email: m.piazza23@campus.unimib.it}}
\author[B]{\fnms{Mauro}~\snm{Passacantando}}
\author[B]{\fnms{Francesca}~\snm{Magli}}
\author[B]{\fnms{Federica}~\snm{Doni}}
\author[B]{\fnms{Andrea}~\snm{Amaduzzi}}
\author[A]{\fnms{Enza}~\snm{Messina}}

\address[A]{Department of Informatics, Systems and Communication, University of Milano-Bicocca, Italy}
\address[B]{Department of Business and Law, University of Milano-Bicocca, Italy}


\begin{abstract}
The interconnected nature of the economic variables influencing a firm's performance makes the prediction of a company's earning trend a challenging task. 
Existing methodologies often rely on simplistic models and financial ratios failing to capture the complexity of interacting influences. 
In this paper, we apply Machine Learning techniques to raw financial statements data taken from AIDA, a Database comprising Italian listed companies' data from 2013 to 2022.

We present a comparative study of different models and following the European AI regulations, we complement our analysis by applying explainability techniques to the proposed models. 
In particular, we propose adopting an eXplainable Artificial Intelligence method based on Game Theory to identify the most sensitive features and make the result more interpretable.
\end{abstract}

\end{frontmatter}


\section{Introduction}

The accurate prediction of a company's future earnings could represent a significant milestone in finance, particularly in accounting research and investment practice. The possible range of applications of such an idea could vary from aiding investment decision-making to evaluating corporate performance \cite{Lev2016}. However, a company's future performance is influenced by numerous concurrent and complex factors and past literature does not provide clear guidance on effective proxies for predicting future earnings~\cite{Lev2016,Monahan2018}. 

Machine Learning (ML) and Deep Learning (DL) models provide the capability to automatically identify patterns and relationships in data, without requiring human experts to have previously identified them. In recent years, these technologies have been applied across various fields to solve disparate problems. Examples include healthcare for disease diagnosis, autonomous vehicles for self-driving cars, natural language processing for sentiment analysis, and computer vision for object recognition. 

Most of the literature regarding profitability prediction relies on data extraction to construct a limited set of financial ratios and simpler linear models, such as Logistic Regression \cite{anand_predicting_2019,belesis_predicting_2023}. 
An exception to this trend is found in \cite{CHEN2022}, where the authors trained a nonlinear classifier on a comprehensive financial statements dataset, achieving better performance than models trained on financial ratios. They conducted a case study on US companies, with XBRL data provided by the SEC organization.

Partially inspired by \cite{CHEN2022}, this work aims to investigate if a similar approach applies to a different context represented by Italian companies; instead of using SEC XBRL we examine Italian data, which are compiled adhering to different regulations. Our dataset comprises listed Italian companies with balance information from 2013 to 2022, sourced from the AIDA Database\footnote{\url{https://login.bvdinfo.com/R0/AidaNeo}}. The experiments are modeled to forecast only the earning direction, without considering its magnitude, because prior research \cite{Gerakos2012} suggests the direction alone is a challenging task, while predicting the magnitude may not be better than the random walk.
We present the results of two comparative studies: the comparison of several classification models based on ML and 
the use of raw data directly extracted from the financial statements compared with the use of financial ratios proposed in~\cite{Ou1989}. 

Additionally, considering the recent European regulation concerning AI and its practical applications, we integrated methods from eXplainable Artificial Intelligence (XAI) to enhance the transparency of ML and DL models. 
To do this, we explore SHAP, a method based on cooperative Game Theory and currently recognized as one of the most famous techniques for XAI.  
The contribution of our work is three-fold:
\begin{itemize}
    \item We investigate the feasibility of predicting the profitability of Italian companies using Financial Statements information. To this end, we construct a Dataset from the AIDA database and compare various classifiers applied to both raw data and financial ratios.
    \item We integrate Game Theory-based methods for XAI into the classification pipeline serving a dual purpose: to justify the model's prediction and provide insights to the auditing literature regarding the most significant features for profitability prediction.
    \item We establish a baseline for this approach and propose future developments to improve the current methodology.
\end{itemize}

\section{Methods: dataset, ML models and XAI approach}
\label{sec:methods}

The dataset object of this study is based on the AIDA Database, provided by the Bureau van Dijk company and contains the financial statements, demographic, and industry data of all active and failed Italian corporations (excluding Banks, Insurance companies, and Public Entities). 
We specifically selected Italian listed companies for which we have 10 years of information (from 2013 to 2022), resulting in a total of 327 companies. For our analysis, we included data from 2013 to 2020 in the training set, while 2021 was used as the test set.
The complete dataset comprised 321 features, divided into four groups: Financial Profile, Balance Sheet, Income Statement and Ratios analysis. Among the total set of features, twenty are excluded for being non-numeric and mainly refer to the company's basic information, such as its name and legal address.

We evaluated various ML models. According to recent literature \cite{ShwartzZiv2022}, tree-based ensembles, particularly XgBoost outperform neural networks in classifying tabular data. 
Xgboost also demonstrated superior performance with financial statements data in \cite{CHEN2022}, hence we chose it as our starting model. 
Additionally, we included in our comparison Random Forest, a different type of tree-based ensemble, a radial-basis kernel Support Vector Machine (SVM), and Logistic Regression. 

We relied on SHAP \cite{lundberg_unified_2017} for the explanation part. 
SHAP, derived from Game Theory, models the features of an ML problem as the players of a cooperative game with transferable utility and estimates the Shapley Value \cite{Shapley1953} on them. The Shapley Value is one of the possible solution concepts available in Game Theory literature. A solution concept is a rule that prescribes how to split the value of the grand coalition among the players of the game. The grand coalition corresponds to the entire set of players. In migrating from Game Theory to Machine Learning the characteristic function is modelled as the difference with a random classifier. Shapley Value corresponds to the average marginal contribution that each player brings to each coalition. The computation of marginal contribution requires the removal of a player from a coalition, however, it is not possible to remove a feature from an already trained ML model. Different solutions are available in the literature to simulate the absence of a feature, among them in this work we implement the substitution with values sampled from a marginal distribution \cite{chen_algorithms_2023}. The exact computation of Shapley Value requires the calculation of marginal contribution for each possible coalition, whose cardinality is equal to $2^n$, where $n$ is the number of players. In most ML problems, including the one object of this study, it is not feasible in a reasonable amount of time, for this reason, a sample is taken from the global set of coalitions.  
This approach is widely used in the field of XAI due to its mathematical properties and the availability of a ready-to-use Python library\footnote{\url{https://shap.readthedocs.io/en/latest/}}.
The name and the framework SHAP hide several approaches and formulations. In this work we employ two of them with different aims: the so-called PartitionSHAP to estimate at a broad level the influence of the four parts composing a document (Financial Profile, Balance Sheet, Income Statement and Ratio Analysis) and the so-called KernelSHAP to estimate the fine-grained importance of each feature. PartitionSHAP estimates the Shapley Value given a fixed partition structure, thus considering some features as always belonging to the same coalition. On the other hand, KernelSHAP is implementing the basic idea of SHAP, modelling each feature as a player and computing the Shapley Value \cite{Shapley1953} on them. 

One of the main open problems in current literature regarding XAI is the absence of a recognized and objective method for evaluating proposed explanations. In this work, to validate the XAI findings, we propose examining how the classification performance changes when using only a subset of features, specifically those with higher SHAP values or excluding those with lower SHAP values.




\section{Experiments and results}

To validate our approach, we conducted a series of experiments to evaluate the model's ability to predict changes in a company's profitability one year ahead. Among the various measures available for representing a company's profitability, we selected the Return On Investment (ROI), a well-recognized metric in the literature \cite{anand_predicting_2019, CHEN2022}. The experiments were designed to compare multiple ML models, ranging from linear models to complex tree ensembles, using two types of data: raw data extracted directly from financial statements and financial ratios proposed in \cite{Ou1989}. To assess the classification performance of the machine learning models, we used two performance measures: accuracy and roc-auc score. Accuracy, the gold-standard measure in ML literature, computes the ratio of correctly classified test elements to the total number of elements in the test set. While accuracy provides a general measure of classifier performance, it can be biased by the class distribution in the test set. To address this limitation and to compare our models with the state-of-the-art, we also included the roc-auc score, which better represents a model's classification performance, especially with respect to its ability to handle the minority class.
 
Once the classification task has been performed, we focused on the integration of Explainability methods.
One of the main problems in the current literature is the absence of a recognized and objective process to assess and evaluate the proposed explanations. Additionally, SHAP is local, meaning it can explain a single element, without providing a broad view of the global decision process followed by the ML model.  
To address these limitations, we designed a series of experiments. Firstly, we applied the Partition-SHAP framework to identify which part of the financial statements has the most significant impact on the final result. Secondly, to provide a more global perspective, we computed the fine-grained SHAP values for each feature and we analyzed the frequency with which each feature appeared among the highest and lowest values. Finally, to validate the results provided by SHAP, we observed changes in classification performance when considering only important features or removing the noisy ones.

\subsection{Classification results}
\label{sec:classification_results}
In this section, computational results regarding classification will be presented.
Table~\ref{tab:results} shows the results 
obtained by different ML models in terms of accuracy and roc-auc score on raw data and financial ratios.
The tree-based models are the best choice for classifying financial statements data, likely due to their strong ability with tabular data. When comparing results on raw data and financial ratios, we note that using high-dimensional raw data is beneficial in representing complex relationships and enhancing final classification accuracy, confirming results obtained in the literature \cite{CHEN2022}. Using financial ratios can simplify the task; however, this approach also omits some fundamental information that simpler linear models cannot fully leverage. Nonetheless, by combining non-linear classifiers and high-dimensional raw data, it is possible to significantly enhance performance.

 


\begin{table}[]
\caption{Comparative analysis of ML models trained on raw financial features versus financial ratios reported in~\cite{Ou1989}.}
\centering
\begin{tabular}{lcccc}
\toprule
                    & \multicolumn{2}{c}{Raw data} & \multicolumn{2}{c}{Financial Ratios} \\
                    & Accuracy         & roc-auc   & Accuracy             & roc-auc       \\ \midrule
XgBoost             & 0.64             &     0.64      & \textbf{0.55}        &        \textbf{0.51}       \\
Random Forest       & \textbf{0.67}    &     \textbf{0.66}      & 0.54                 &       0.50        \\
SVM                 & 0.57             &      0.57     & 0.54                 &      0.50         \\
Logistic Regression & 0.57             &    0.54       & 0.54                 &        0.50       \\ \bottomrule
\end{tabular}
\label{tab:results}
\end{table}

\subsection{XAI results}

\begin{figure}[htb]
\centering
\includegraphics[width=0.45\textwidth]{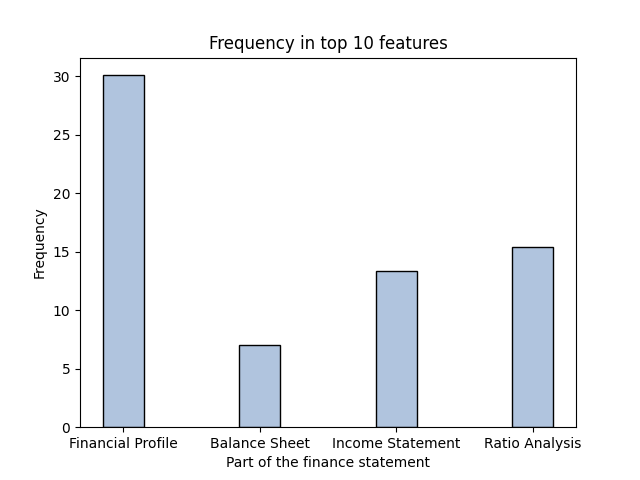}
\caption{Frequency in Top 10 ranked features, normalized by the number of features composing a specific financial statement' part.}
\label{fig:barplot}
\end{figure}

In this section, we present and briefly discuss the results obtained with eXplainable Artificial Intelligence. 
Firstly, by using the PartitionSHAP framework, we computed the results aggregating the features belonging to the four different parts of the financial documents: Financial Profile, Balance Sheet, Income Statement and Financial Ratios. In fact, due to the considerable number of features, capturing the intricate relationship between each of them becomes challenging. However, the obtained results were comparable between each document's parts.
For this reason, to deepen the analysis we examine the probability of each feature being selected among the Top Ten features with the highest SHAP values. Figure \ref{fig:barplot} shows the normalized frequency of features appearing in the \textit{"Top-10"}. As illustrated, features related to the financial profile are consistently selected as the most important. This part summarizes the content of financial statements with a limited set of measures and financial ratios. The XAI analysis suggests that they are, on average, the best predictors of profitability. However, using these features alone is not sufficient for robust classification, as demonstrated by the classification results in Section \ref{sec:classification_results}.

\begin{figure}[htb]
\centering
\includegraphics[width=\columnwidth]{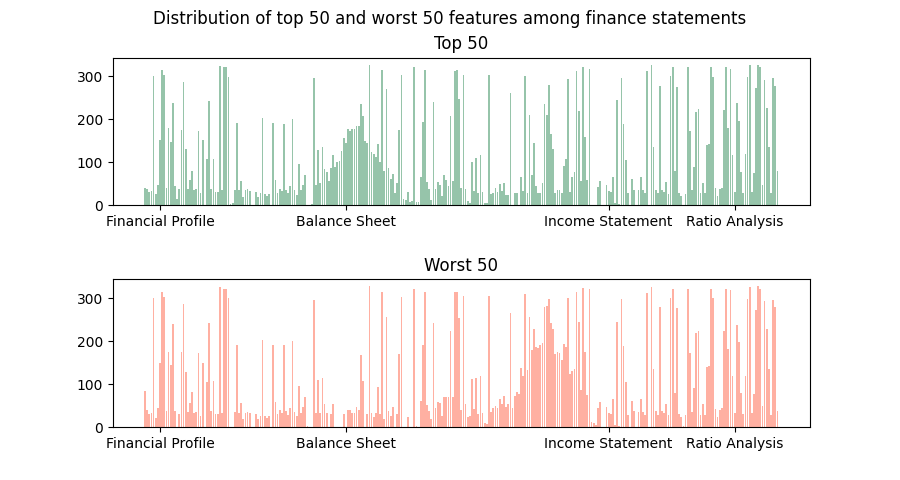}
\caption{Comparative analysis of the distribution along the financial statements of the 50 most important and 50 less important features. The importance metric is computed with KernelShap.}
\label{fig:shap_distribution}
\end{figure}

To complement the XAI results we analyze how the features recognized as important are distributed along the financial statements. This is achieved by selecting the fifty most important (\textit{"Top-50"}) and least important (\textit{"Worst-50"}) features for each element of the test set. The resulting distribution is depicted in Figure \ref{fig:shap_distribution}. 

Each part of the financial statements contains some important features, highlighted by the presence of peaks in each of them in the first panel of Figure \ref{fig:shap_distribution}.
It is noteworthy that the features between the latter part of the Balance Sheet and the initial part of the Income Statement are useless for the classification. The distribution of features across the Financial Statements also confirmed the prevalent importance of features related to Financial Profile and Ratio Analysis. To complete the analysis, the names of the features corresponding to higher peaks in the distribution plot were identified. Features constantly among those with higher SHAP value are mainly related to debt, equity and taxes, while features constantly among those with lowest SHAP value are minor entries of the financial statements, which often can also be unfilled.

\subsection{XAI Validation}

The validation of XAI results is carried out by evaluating the performance of classification models on subsets of features identified as important according to their SHAP values and comparing this performance with those of models trained on the entire set of features. Firstly, the features are ranked according to the number of times they appear in the Top-50 list. Secondly, the top and bottom hundred features from this ranking are selected. Finally, one classification is performed considering only the top hundred features, while the other is performed considering the total number of features, excluding the bottom hundred, resulting in 201 features. The experiments are performed using an XgBoost classifier.


When classifying using only Top-100 features, there is a slight decrease in classification performance, as shown in Table \ref{tab:validation_one}. However, the removal of such a high portion of features does not significantly impact the classification ability. On the other hand, eliminating less important features leads to an increase in classification performance, up to 0.04 accuracy points.

\begin{table}[htb]
\centering
\begin{tabular}{@{}lc@{}}
\toprule
                 & \multicolumn{1}{l}{Accuracy} \\ \midrule
All features     & 0.64                         \\
Top-100 Features & 0.61                         \\
Top-201 Features & \textbf{0.68}                \\ \bottomrule
\end{tabular}
\caption{Classification accuracy obtained performing the validation of XAI approach, compared with classification accuracy on the entire set of features.}
\label{tab:validation_one}
\end{table}

SHAP computes the importance value of each element relative to each class. Therefore, we investigated whether there are differences in features identified with respect to the class corresponding to a decrease in ROI (modelled as 0) and the class corresponding to an increase in ROI (modelled as 1). To do this, we replicated the same analysis described above for each class. The best results were obtained using the Top-201 features regarding class 0 (0.69 accuracy). Similarly, in the case of the Top-100 features, those identified from class 0 led to higher accuracy (0.64). Using features identified with respect to a single class slightly improves the classification ability; nevertheless, the increment is not significant.   

This analysis confirms that SHAP is partially effective in identifying the most influential features for classification and can reduce the dimensionality, without losing classification ability. It is also more performant in identifying redundant and noisy features that can be removed from the dataset.

\section{Conclusions and Future works}

In this work, we presented the idea of leveraging the power of ML models to automatically explore the information contained in listed companies' financial statements for profitability prediction. 
Although several works in the literature have addressed this task~\cite{anand_predicting_2019, belesis_predicting_2023}, only \cite{CHEN2022} used raw high-dimensional features for this purpose. Partially inspired by the approach proposed in \cite{CHEN2022}, we investigated its applicability in a different context represented by Italian listed companies. Additionally, to fulfil the recent AI regulations requirements from the European Union, we designed the classification pipeline considering also an XAI approach, based on the popular Game Theory framework SHAP \cite{lundberg_unified_2017}.

As demonstrated by the aforementioned results it is possible to predict the one-year earning change of a company by combining ML techniques with tabular information extracted from financial statements, obtaining performance consistent with the state-of-the-art \cite{CHEN2022} or even slightly better \cite{anand_predicting_2019, belesis_predicting_2023}. However, it is worth noting that the dataset proposed for this study contains less information and is composed of a smaller time horizon with respect to those used in past works.
Comparative analysis of various ML models validates previous findings, indicating tree-based ensembles as the optimal choice for tabular data processing. Moreover, this analysis confirmed that high dimensional data offer richer information compared to simpler financial ratios leading to improved classification performance. 

Regarding eXplainable Artificial Intelligence, the assessment of feature influence was first studied by aggregating the features composing a financial statement, namely: Financial Profile, Balance Sheet, Income Statement and Ratios Analysis. 
This analysis showed that each of these parts has a comparable influence on the final classification. 
By deepening the XAI analysis was possible to demonstrate that important features are distributed among all the financial statements, but most of them were identified among the financial profile. In order to objectively evaluate the features identified with SHAP we compare the classification performance on the entire dataset and two reduced sets of features, demonstrating that the information provided by SHAP is able to identify the more sensitive and more noisy features.


The project is still in its early stages, with our primary objective being the presentation of the idea and the examination of its feasibility. For this reason, it is worth discussing potential future developments that could enhance the current methodology.

To enhance the classification performance we plan to extend the model in multi-modal settings, integrating the free text information contained in additional notes. In this scenario, the manipulation of textual information could not ignore the use of DL models and particularly recent powerful Large Language Models. 
Additionally, Machine Learning is well-known to be a data-hungry technology as the training of automatic algorithms requires a lot of data, for this reason, the augmentation of the current dataset with more years, companies and markets can be fundamental to fully explore the potential of ML-based algorithms.  

\begin{ack}
This work was partially supported by the MUR under the grant “Dipartimenti di Eccellenza 2023-2027" of the Department of Informatics, Systems and Communication of the University of Milano-Bicocca, Italy.
\end{ack}


\bibliography{main}

\end{document}